%% file: ms.tex
\documentclass[journal]{IEEEtran}

\IEEEoverridecommandlockouts                              

\usepackage{cite}
\usepackage{amsmath,amssymb,amsfonts}
\usepackage{graphicx}
\usepackage{textcomp}

\usepackage{stfloats}
\usepackage{amsmath}
\usepackage{graphicx}
\usepackage{amsxtra}
\usepackage{subfigure}
\usepackage{ifpdf}
\usepackage{booktabs}
\usepackage{amssymb}
\usepackage{enumerate}

\usepackage{flushend}

\usepackage{pgfplotstable,filecontents}
\pgfplotsset{compat=1.9}

\usepackage{amsmath}
\usepackage{multirow}
\usepackage{stfloats}
\usepackage{algpseudocode}
\usepackage{epsfig}
\usepackage{tensor}
\usepackage[linesnumbered,ruled,vlined]{algorithm2e}
\usepackage{printlen}
\usepackage{commath}

\usepackage{todonotes}

\usepackage{comment}

\title{\LARGE \bf Federated Reinforcement Learning for Collective Navigation of Robotic Swarms}

\author{Seongin Na$^1$, Tom\'{a}\v{s} Rou\v{c}ek$^2$, Ji\v{r}\'{i} Ulrich$^2$, Jan Pikman$^2$, Tom\'{a}\v{s} Krajn\'{i}k$^2$, Barry Lennox$^1$ and Farshad Arvin$^3$
\thanks{This work was supported by EU H2020-FET RoboRoyale [964492] and Czech Science Foundation (GA\v CR) through 20-27034J.  }
	\thanks{$^1$S. Na and B. Lennox are with Department of Electrical and Electronic Engineering, The University of Manchester,  Manchester, M13 9PL, UK. \\(e-mail: seongin.na@manchester.ac.uk). \\ $^2$T. Rou\v{c}ek, J. Ulrich, J. Pikman and T. Krajn\'{i}k are with Czech Technical University in Prague, FEE, AIC%
	\\
	$^3$F. Arvin is with the Swarm \& Computational Intelligence Laboratory (SwaCIL), Department of Computer Science, Durham University, UK.\\ (e-mail: farshad.arvin@durham.ac.uk)}
}

\begin{document}
\maketitle
\thispagestyle{empty}
\pagestyle{empty}

\begin{abstract}
The recent advancement of Deep Reinforcement Learning (DRL) contributed to robotics by allowing automatic controller design. The automatic controller design is a crucial approach for designing swarm robotic systems, which require more complex controllers than a single robot system to lead a desired collective behaviour. Although the DRL-based controller design method showed its effectiveness, the reliance on the central training server is a critical problem in real-world environments where robot-server communication is unstable or limited. We propose a novel Federated Learning (FL) based DRL training strategy (FLDDPG) for use in swarm robotic applications. Through the comparison with baseline strategies under a limited communication bandwidth scenario, it is shown that the FLDDPG method resulted in higher robustness and generalisation ability into a different environment and real robots, while the baseline strategies suffer from the limitation of communication bandwidth. This result suggests that the proposed method can benefit swarm robotic systems operating in environments with limited communication bandwidth, e.g., in high-radiation, underwater, or subterranean environments.
\end{abstract}
\begin{IEEEkeywords}
Federated Learning, Deep Reinforcement Learning, Swarm Robotics, Collective Navigation.
\end{IEEEkeywords}


\input{src/introduction}

\input{src/methodology}
\input{src/experiments}

\input{src/results}

\input{src/conclusion}

\section*{Acknowledgements}
Zdeněk Rozsypálek from Chronorobotics laboratory in Czech Technical University provided important discussions and insights on Deep Reinforcement Learning algorithm training and real robot experiments.




\bibliographystyle{IEEEtran}
\bibliography{IEEEabrv, bibfile}

\end{document}

%% file: src/introduction.tex
\section{INTRODUCTION}

Swarm robotics is a field of study of how a large number of robots can coordinate for a common goal with local communication and decentralised control \cite{schranz2021}. 
It is inspired by social insects that form swarms, such as ants that perform long-distance foraging using pheromones. 
As social insects collectively conduct challenging tasks, which are impossible for single individuals, swarm robotic systems are prospective for the challenging missions under dynamic and complex environments, which are difficult for single robots \cite{Brambilla2013}. 
Swarm robotic systems have already been implemented for such missions or are expected to be developed in future. 
For example, a swarm of unmanned surface vehicles (USV) for maritime environmental monitoring and border patrolling \cite{Duarte2016}. 
Furthermore, more diverse applications are expected in future, such as planetary exploration with miniature robot swarms, targeted drug delivery with microscopic robot swarms \cite{Dorigo2020futureOfSwarm} and agri-robotic applications~\cite{GRIEVE2019agri}. 
To enable the aforementioned applications, recent research works tackled several issues for swarm robotics that can occur in real-robot deployments. 
For example, a self-configured multi legged robot swarm have shown its potential in the terradynamically challenging environment \cite{Ozkan-Aydin2021SciRob} and a swarm robotic system with block-chain technology demonstrated improved security of swarm robotic communication \cite{Ferrer2021Blockchain}. 
 
While the future of swarm robotics seems bright, there are  significant technical challenges  to be overcome for further development of swarm robotic systems \cite{Brambilla2013, Francesca2016Automatic}. 
One of the significant challenges is how to design the individual behaviour of a robot in a swarm. 
As swarm robotic systems do not use a centralised unit to control the entire swarm, careful design of individual controllers is essential to achieve desired swarm behaviours \cite{Hamann2018, sahin2008}. 
The current most dominant strategy to design the behaviour of individual robots is the manual design method \cite{Brambilla2013}. 
It derives individual behaviour rules heuristically from the target swarm behaviour and finds an optimal controller through iterative tuning. 
This strategy is suitable for simple scenarios. However, it is highly challenging to manually derive individual controllers for more complicated tasks and complex environments. 
Furthermore,  manually derived controllers cannot adapt to environmental change \cite{Birattari2019Automatic}.

\begin{figure*}[t]
    \centering
    \includegraphics[width=0.98\textwidth]{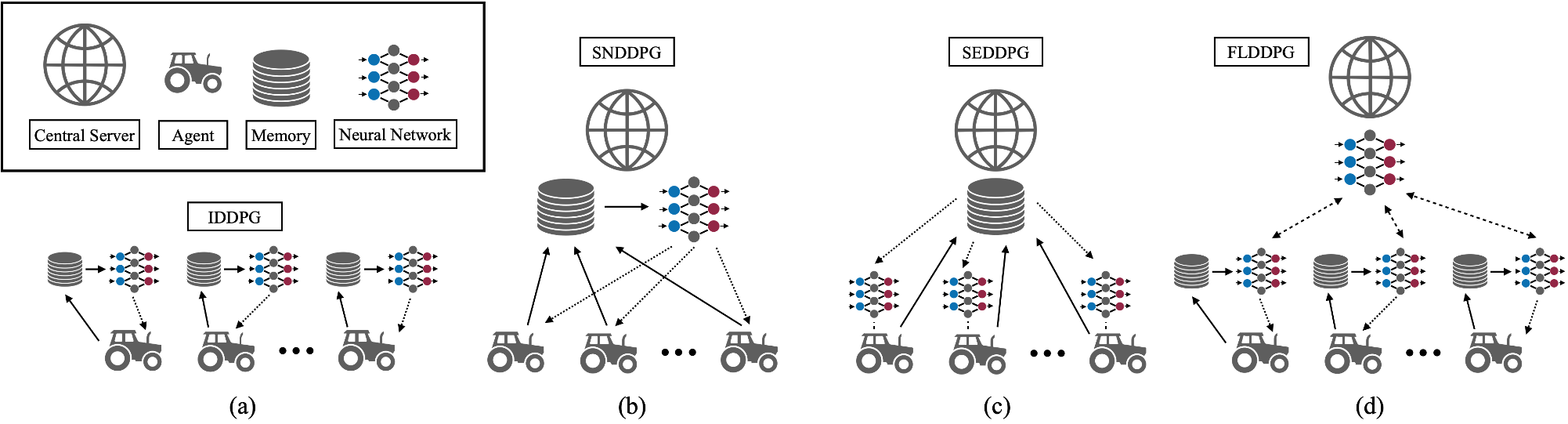}
    \caption{Deep reinforcement learning training architectures for swarm robotic systems. (a) Individual DDPG (IDDPG), (b) Shared Network DDPG (SNDDPG), (c) Shared Experience DDPG (SEDDPG) and (d) proposed Federated Learning DDPG (FLDDPG). }
    \label{fig:algorithms}
\end{figure*}

In contrast to manual design method, automatic design method gives ability to individual robots to find the optimal behaviour autonomously without external intervention.
One of the automatic design methods to design swarm robotic controller is Reinforcement Learning (RL)~\cite{Sutton2018}.
RL allows to find an optimal controller from interactions with the environment in a way that it maximises the total reward in a given mission period.
Recently, Deep Reinforcement Learning (DRL), which uses deep neural networks as a function approximator, has been rapidly developed and showed the potential in diverse domains, e.g. multi-robot systems~\cite{2020HuTVT, Na2022TVT, Chen2017a, Liu2022DNN} and self-driving cars~\cite{kiran2021deep}.
The effectiveness of DRL for swarm robotic system has been demonstrated e.g. in \cite{Fan2020TensorSwarm}, where DRL considerably outperformed manual controller design in the navigation and collision avoidance tasks with up to 100 robots.
To overcome the limitation of manual design, researchers proposed and developed the automatic design method~\cite{Birattari2019Automatic}. 
In contrast to the manual design method, the automatic design method allows individual robots to find the optimal behaviour autonomously without external intervention. 
One of the automatic design methods to synthesise a swarm robotic controller is Reinforcement Learning (RL)~\cite{Sutton2018}. 
RL allows finding an optimal controller from interactions with the environment to maximise the total reward in a given mission period. 
Recently, Deep Reinforcement Learning (DRL), which uses deep neural networks as a function approximator, has been rapidly developed and showed potential in diverse domains, e.g. multi-robot systems~\cite{2020HuTVT} and self-driving cars~\cite{kiran2021deep}. 
Besides other applications, DRL's effectiveness for swarm robotic systems has been demonstrated in navigation scenarios \cite{Fan2020TensorSwarm}, where DRL considerably outperformed manual controller design.
.
 
Although the feasibility and benefits of RL have been demonstrated and expected to produce more promising results as it develops, there is a critical issue in implementing RL for swarm robotic systems in real-world applications.
In the simulated environments, all data collected from individuals are used for training in a central server \cite{Huang2020, Fan2020TensorSwarm, na2021universal}.
The dependence on a central server is present in Multi-Agent Reinforcement Learning (MARL), which is a sub-domain of RL dealing with multi-agent scenarios.
In several MARL research works, centralised training and decentralised execution scheme is used for training multi-agent systems, e.g., COMA~\cite{foerster2017counterfactual} and MADDPG~\cite{lowe2020multiagent}.
Although the controllers used for each agent can differ, training is conducted in the centralised server.
However, perfect and high-bandwidth communication between large numbers of robots and a central server cannot be ensured in the real world.
Communication constraints apply to scenarios where robotic swarms would be especially effective, e.g., underwater inspection, extreme environment exploration, or subterranean search and rescue~\cite{Roucek2020DARPA}.
In the conditions of limited communication, a centralised, data-intensive way of training swarm robotic systems for real-world applications can become ineffective or even impossible.
Therefore, DRL training strategy that reduces communication bandwidth between a large number of robots and the central server is necessary.

Federated learning (FL) is an emerging paradigm for distributed training of machine learning models with a large number of agents~\cite{BrendanMcMahan2017FL}. 
In an FL setting, each agent trains an individual model using locally collected data. 
The individually trained models are then periodically shared with the central server that aggregates them and transmits a refined model back to the individual agents. 
Since it does not share the sensorimotor data between the agents, it is used as a learning paradigm for domains where security and privacy are  a concern, e.g., secure 5G networks~\cite{Liu2020FL5G} and multi-institutional medical diagnosis~\cite{Sheller2020FLmedicine}. 
This paradigm has three significant advantages for swarm robotic scenarios: i) major reduction in communication with the server, ii) privacy preservation and iii) customisation of models. 
First, as each individual only communicates with the server to share the model, the volume of exchanged data can be considerably reduced compared to the conventional training methods. 
Second, since it uses local sensorimotor data to train the local model and sends only the model, the local sensory data remain concealed within the agent. 
This avoids security and privacy issues when deploying a swarm robotic system in strategic infrastructure facilities or public areas. 
Finally, as the individual model is trained using the local data, it can better adapt to the individual characteristics of an actual agent. 
In the real world,  swarm robots are not identical, but they differ slightly due to the nature of their manufacturing and wear and tear during their operation. 
With the great potential of integration of FL with DRL, FL-based DRL has been validated in a simple grid-world environment \cite{zhuo2020federatedDRL} and internet-of-things applications~\cite{Wang2020FDRLIOT}. 
In a robotic application, it showed its advantages in navigation and collision avoidance scenarios~\cite{Liu_2019FRL}. 
However, the authors of~\cite{Liu_2019FRL} used only a single robot in different environments and trained it sequentially.

This paper proposes an FL-based DRL training strategy for swarm robotic systems (FLDDPG).
We evaluate four DRL training strategies, including the FLDDPG, in both simulated and real-world scenarios, where a swarm of robots collectively learns to navigate towards given destinations while avoiding obstacles. 
We focus our evaluation on the performance of these strategies under the limited communication bandwidth, which reflects a major challenge of real-world swarm robotic missions.
The obtained results from the experiments demonstrated that FLDDPG resulted in higher robustness and generalisation ability compared to the baseline strategies.

%% file: src/methodology.tex
\section{METHODOLOGY}

This section introduces the proposed federated learning (FL)-based DRL training strategy for swarm robotic systems.
First, we describe design choices when applying DRL for swarm robotic systems.
Second, we explain the selected DRL algorithm.
Third, we describe three traditional DRL training strategies.
Finally, we illustrate the proposed FL-based DRL training strategy for swarm robotic systems.
Figure~\ref{fig:algorithms} illustrates how the three traditional training strategies and our proposed strategy are utilised for swarm robotic systems in this study.

\subsection{DRL Design Choices for Swarm Robotic Systems}

DRL training for swarm robotic systems requires two design choices: i) training algorithm and ii) training strategy. 
Regarding the training algorithm, the question is what particular DRL algorithm is suitable for the given agent and environment. 
For example, when continuous action is required for the agent, e.g. mobile robots \cite{Fan2020TensorSwarm, 2020HuTVT}, DRL algorithm that outputs continuous values needs to be chosen, e.g., the DDPG \cite{lillicrap2019continuous}, PPO~ \cite{Schulman2017PPO} and TD3~\cite{fujimoto2018TD3}. 
In contrast, when discrete actions are required, e.g. robot moves in a grid-like environment, DRL algorithm for discrete action needs to be used, e.g., DQN \cite{Mnih2015}.

Training strategy refers to how to train multiple members of a robot swarm using the selected DRL algorithm.
Two factors need to be considered when choosing a suitable training strategy.
First, whether the memory for collected transition samples is shared or not, i.e.,  local memory vs shared memory.
Second,  whether neural network training is conducted in the central server or not, i.e., centralised training vs decentralised training.

\subsection{DRL Training Algorithm}

We selected Deep Deterministic Policy Gradient (DDPG) algorithm \cite{lillicrap2019continuous} as the training algorithm for this work.
DDPG is an actor-critic, model-free DRL algorithm for agents with continuous observation and action spaces.
It is one of the successful DRL algorithms in continuous observation and action domains, e.g., mobile robots \cite{2020HuTVT, Fan2020TensorSwarm}, mobile edge computing \cite{Huang2020} and unmanned aerial vehicles \cite{Liu2018UAV}.
Since robots in a swarm require continuous observations and actions as the described application domains, we chose DDPG to train individual robot in a robot swarm.
We recommend the readers to check the original paper for the technical details of DDPG \cite{lillicrap2019continuous}.

DDPG algorithm has a important feature that contribute to applications with multiple agents as well as single agent scenarios: experience replay.
The agent collects transition samples into an experience replay memory and the samples are selected from the memory when the neural network is trained.
As the memory can be accessed by more than one agent, this algorithm can be used for multi-agent scenarios.

\subsection{Traditional DRL Training Strategies}
Here we introduce three traditional DRL training strategies using DDPG algorithms: i) Independent DDPG, ii) Shared Network DDPG and iii) Shared Experience DDPG.

\subsubsection{Independent DDPG}
Independent DDPG (IDDPG) training strategy is the most na\"ive application of DDPG to train a robot swarm. Fig. \ref{fig:algorithms}~(a) illustrates IDDPG training strategy. IDDPG deploys individual neural networks and local memory for each robot in a swarm. In other words, there is no communications between the robots and the server for training. The advantage of IDDPG is that it does not use a central server to share data samples or neural network parameters. However, this training strategy does not allow to utilise the collective nature of swarm robotic systems for  learning.

\subsubsection{Shared Experience DDPG}
Shared Experience DDPG (SEDDPG) training strategy is one of the state-of-the-art training strategy for multi-agent systems \cite{Christianos2020SEAC}. Fig. \ref{fig:algorithms}~(c) illustrates SEDDPG training strategy. In this training strategy, the agents have individual neural networks and a shared memory. In the original paper, it is stated that the advantage of SEDDPG is that by sharing it can encourage exploration, thereby faster convergence and better performance. Despite the improvement in the training speed and performance, the robots still share the collected data with the central server, which requires significant communication bandwidth. 

\subsubsection{Shared Network DDPG}
Shared Network DDPG (SNDDPG) is a training strategy that uses a shared neural network and a shared memory. Fig. \ref{fig:algorithms}~(b) illustrates SNDDPG training strategy. In each time step, the robots transfer the transition sample data to the shared memory located in the central server. In every training period, the network update is performed by the central server and the model is distributed to the individual agents. The advantage of this method that the central server can be trained using the data samples collected by different agents in different environments. Compared to IDDPG, it encourages the networks to learn from more diverse data, leading to a more generalised controller. However, it requires frequent communication between each robot and the central server. The more frequent the communication between robots and the central server is required, the more vulnerable the training is as the communication is unstable.




\subsection{Federated DRL Training Strategy}
\begin{algorithm}[t]
\SetAlgoLined
Randomly initialise $N$ critic neural networks, $Q_{1, ... , N}(s, a|\theta^Q_{1, ... , N})$ and $N$ actor neural networks $\pi_{1, ... , N}(s|\theta_{1, ... , N}^{\pi})$ with weights $\theta_{1, ... , N}^{Q}$ and $\theta_{1, ... , N}^{\pi}$\\
Initialise $N$ target networks $Q_{1, ... , N}^\prime$ and $\pi_{1, ... , N}^\prime$ with weights $\theta_{1, ... , N}^{Q^\prime}$ $\leftarrow$ $\theta_{1, ... , N}^Q$, $\theta_{1, ... , N}^{\pi^\prime}$ $\leftarrow$ $\theta_{1, ... , N}^{\pi}$\\
Initialise replay buffers $R_{1, ... , N}$\\
\For{episode = 1, $...$, M}
{ Initialise the states $s_t = s_1$\\
\For{t = 1, $...$, T}{
Run $N$ actors and collect transition samples $D_t$ = ($s_t, a_t, r_t, s_{t+1}$) into $R_{1, ... , N}$ \\

  \If{t = 0 mod $t_{train}$}
  {Sample $l$ transitions ($s_i, a_i, r_i, s_{i+1}$) from local replay buffer memories, $R_{1, ... , N}$\\
  Set $y_i = r_i + \gamma Q^{\prime}(s_i, \pi^{\prime}(s_{i+1}|\theta^{\pi^\prime})|\theta^{Q^\prime})$\\

  Update critic networks by minimizing the loss: 
  $L_Q = \frac{1}{N}\sum_i(y_i-Q(s_i,a_i|\theta^Q))^2$\\
  Update actor networks using the sampled policy gradient:

  $\nabla_{\theta^\pi}J \approx \frac{1}{N}\sum_i \nabla_a Q(s,a|\theta^{Q})|_{s=s_i, a=\pi(s_i)} $\par $\quad\quad\quad\quad \cdot\nabla_{\theta^\pi}\pi(s_i|\theta^\pi)|_{s=s_i}$\\ 
  }
  \If {t = 0 mod $t_{target}$}{Update the target networks:
  
  $\theta^{Q^\prime} \leftarrow \theta^Q$,
  $ \theta^{\pi^\prime} \leftarrow \theta^{\pi}$}
 }
 \If {episode = 0 mod $T_{wa}$}{Perform \textit{Soft Weight Update} for actor and critic networks for all robots \\ 
 
 $\theta_{wa} = \frac{1}{N}\sum_{k=1}^{N} \theta_k$ \\

\For{i = 1, $...$ , N}
{$\theta_i = \tau\theta_i + (1-\tau)\theta_{wa}$}
 }
	}
	
\caption{DDPG with FL (FLDDPG)}
\label{algo:FLDDPG}
\end{algorithm}

Using the concept of FL and DDPG algorithm, we proposed a FL based DRL training strategy, FLDDPG. The process of FLDDPG is illustrated in Fig. \ref{fig:algorithms}~(d), showing only the neural network weights are shared in the central training server, without sharing the locally collected data. This feature can largely reduce the number of communication between the robots and the server as collected data sharing takes a large part in DRL setting as in SNDDPG and SEDDPG.
 
 Algorithm~\ref{algo:FLDDPG}~(d) describes the algorithm of FLDDPG. At the beginning of the algorithm, actor and critic networks and local memories are initialised with the number of robots. The data collection and neural network parameter update are performed individually for each robot. After initialisation, the transition samples are collected from the robots for $T$ time-steps and stored in the replay buffer in every episode. Every $t_{train}$ times in every episode, the actor and critic networks are updated by the following process. First, $l$ number of transition samples are randomly selected from the local memories. Second, the target, $y_i$, is calculated for each transition sample.
Third, using the target, the temporal-difference loss, $L_Q$, is calculated and the critic networks are updated in the way minimising the loss. Then, the actor networks are updated using the sampled policy gradient. Additionally, for every $t_{target}$ time steps, the newest actor and critic networks are assigned to the target networks.  In Algorithm~ \ref{algo:FLDDPG}, lines 19 to 24, we describe the process of neural network weight averaging and update in the central server. In every weight averaging and update period, $T_{wa}$, the local neural network weights are updated with the averaged weights. For more efficient weight update, we proposed a new method called \textit{Soft Weight Update}. In the seminal work of FL \cite{BrendanMcMahan2017FL}, the hard weight averaging called \texttt{FedAvg} was performed to average the local neural network parameters. The hard weight averaging is described in (\ref{eq:hwa}).
\begin{equation}
\centering
\begin{aligned}
    \theta_{wa} = \frac{1}{N}\sum_{k=1}^{N} \theta_k \\
    \theta_{1, ..., N} = \theta_{wa},
    \label{eq:hwa}
\end{aligned}
\end{equation}
where $\theta_{wa}$ denotes the averaged weights of neural network, $N$ is the number of robots, and $\theta_{1,...N}$ are neural network weights for $N$ robots. With the hard weight update method, the averaged weights are directly assigned to the local neural network weights. The problem of hard weight update method is that whenever the local weights are updated with the averaged weights, adverse change in neural network can occur, decreasing the efficiency of individual controllers in their corresponding environments and tasks after the update. To prevent such adverse changes in the neural network update after the weight update, we proposed a soft weight update method. The concept of soft weight update is that the local neural network weights are fractionally updated with the averaged weights. In the line 23 of Algorithm~\ref{algo:FLDDPG}, the neural network weights are updated with the sum of $\tau\theta$ and $(1-\tau)\theta_{wa}$, where $\tau$ represents an update constant within the range $[0, 1]$. When $\tau$ is close to zero, the local neural network weights are completely replaced with the averaged ones. In contrast, when $\tau$ is close to 1, it becomes analogous to IDDPG.

%% file: src/experiments.tex
\section{EXPERIMENTS}

\begin{figure}[t]
    \centering
    \includegraphics[width=0.48\textwidth]{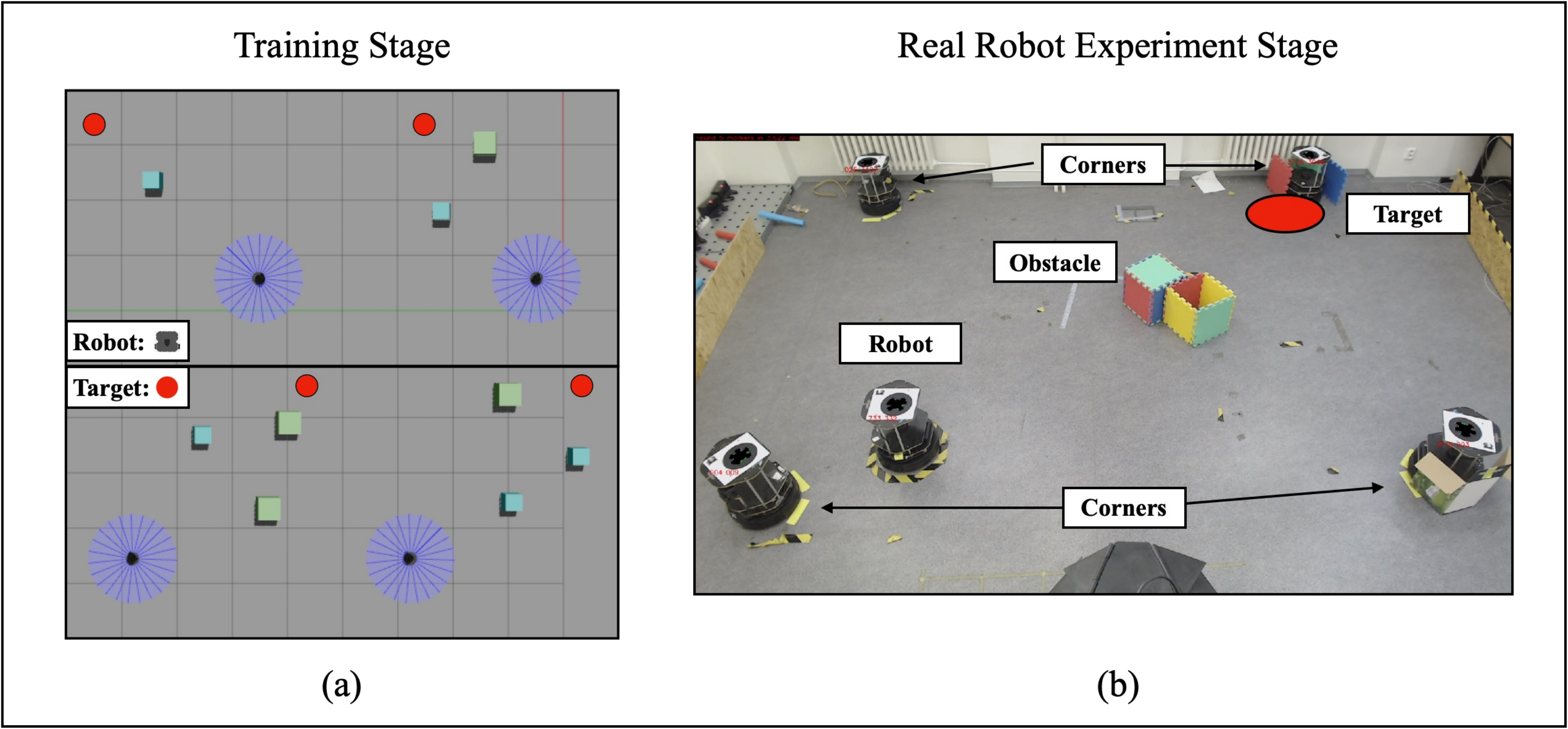}
    \caption{Experimental arenas for (a) training and (b) for real-robot experiments. In (a), four robots learn navigation to the target and collision avoidance independently under different environmental configurations. In (b), a robot performs navigation to the target and collision avoidance in a different environment than the training environment.}
    \label{fig:experiment_environment}
\end{figure}

We designed the experiments to evaluate the performance of
four different training strategies, IDDPG, SEDDPG, SNDDPG
and FLDDPG, under the limited communication bandwidth
scenario. To evaluate the training strategies with the swarm
robotic scenario, we designed a collective learning scenario of
navigation and collision avoidance for swarm robotic systems.
  Fig.~\ref{fig:experiment_environment} illustrates the collective learning simulation environment and its real-robot evaluation environment. In Fig.~\ref{fig:experiment_environment} (a), the robots learn navigation and collision avoidance using the four training strategies. Although the robots do not communicate each other for collective behaviours, the robots help other robots to obtain a better controller utilising diversity of collected samples and trained models for larger number of agents than a single robot. Therefore, the collective learning scenario can be regarded as a swarm robotic scenario.
  
  After training models, evaluation of the trained models were performed in the same environment in Fig.~\ref{fig:experiment_environment} (a). To evaluate
the robustness and generalisation ability of the trained models
with the four strategies, the real-robot platform was deployed
in the environment illustrated in  Fig.~\ref{fig:experiment_environment} (a). The trained robot
used in the simulated environment is the Turtlebot3 burger robot,
which is a differential drive mobile robot with laser sensors
to detect obstacles. The real-robot platform used for the real
robot experiment is described in the later section. In the real-robot experiment, only a single robot is used to assess clearly
the performance of a collectively learned model from the swarm
robotic systems rather than deploying multiple robots, which
can add a higher degree of complexity affecting the evaluation of
the trained model.
  
To add the constraint of limited communication bandwidth,
we calculated the total data volume, a product of
communication bandwidth and total communication time. We
calculated the total data volume from the total data transferred
during the training with FLDDPG. For each transfer of neural
network parameters, it takes 0.55 MB one-way and 1.1 MB
for the complete cycle of weight update. Since the weight update period,  $T_{wa}$ = 1, 120 weight update occurs for one
training instance. Therefore, the total transferred data volume
for training with FLDDPG is 132 MB. 

With this total data volume, the update period for SEDDPG and SNDDPG were chosen to distribute update events over the training evenly.
Both SEDDPG and SNDDPG include the data transfer of
experience replay buffer, which size is 2.4 MB for one-way
and 4.8 MB for the full cycle of transfer. For SEDDPG, the update
period of 5 was chosen over the total 120 episodes as the
total transferred data volume is 115.2 MB, while when the
update period is four it is 144 MB, which is greater than the
upper limit. For SNDDPG, a total 2.95 MB is transferred per
update. Therefore, the update period of 3 was chosen over the
total 120 episodes, resulting total transferred data volume of
118 MB. Unlike the three algorithms, the limitation of the total transferred data volume does not affect training process of IDDPG. The training of models using four strategies applied
to these update period settings.

\subsection{DRL Implementation}
Here we provide the DRL implementation details for four training strategies. The four important design specifications are introduced as: i) observation space, ii) action space, iii) reward design and iv) actor and critic neural network structure.
 
\subsubsection{Observation Space}

During the training, the robot collected observation to learn navigation and collision avoidance. For navigation, the distance between the target and the current position of the robot in polar coordinates ($d, \theta_d$) were collected. For collision avoidance, 24 sensor readings from the laser rangefinder were collected. After collecting the 24 laser sensor readings, the readings are normalised in the range between [0, 1], and the normalised readings are inversed so that when the obstacle is close to the robot, the normalised sensor reading is close to 1 for a more effective neural network training. While the range of the laser sensor is up to 3.5~m, only the value below 0.8~m was used to learn more effective collision avoidance algorithm.

\subsubsection{Action Space}
There are two action values in our setting: i) translational velocity, $v$, and ii) rotational velocity, $\omega$, i.e. $a = [v, \omega]$.  The range of velocities were limited to $v~\in~(0, 0.25)~$m/s and $w~\in~(-\frac{\pi}{2}, \frac{\pi}{2})$~rad/s in translation and rotation respectively to reflect the motion constraints and safety for the robot.

\subsubsection{Reward Design}
There were three types of reward functions that were applied in the experiments. The reward functions were designed to enable the learning of i) navigation, ii) collision avoidance. Equation~(\ref{eq:reward function}) describes the reward functions used in the experiments.

\begin{equation}
\begin{aligned}
& r = r_g + r_p + r_c + r_a\\
& r_g = \begin{cases}
    R_g, & \text{if arrived goal}\\
    0, & \text{otherwise}
  \end{cases},\\
& r_p = \begin{cases}
    ad, & \text{$d$ $>$ 0}\\
    -ad, & \text{otherwise}
  \end{cases}, \\
& r_c = \begin{cases}
    R_c, & \text{if collision}\\
    0, & \text{otherwise}
  \end{cases},\\ 
& r_a = \begin{cases}

    -e^{max(s_{laser})*\lambda}, & \text{any($s_{laser}$) $>$ 0}\\
    0, & \text{otherwise}
  \end{cases}
  \\
    \label{eq:reward function}
\end{aligned}
\end{equation}
The total reward, $r$, was the sum of a series of rewards, $r_g$, $r_p$, $r_c$ and $r_a$. In $r_g$, the goal arrival reward, $R_g$, is given when the robot arrived goal. In $r_p$, $ad$ or $-ad$ is given depending on whether the robot is approaching to or moving away from the goal (specified to be equal to a step length, $d$, multiplied by a tunable factor, $a$, which was set empirically to 4.0), In $r_c$, the collision penalty, $R_c$ is given when the robot collides to the obstacle. In $r_a$, the approaching penalty is given when any of the laser sensor value, $s_{laser}$, is higher than zero, i.e., the laser finder detects obstacles. $\lambda$ is a parameter to set the intensity of approaching penalty. The parameter values used in the experiments are provided in Table \ref{table:parameters}.
\subsubsection{Neural Network Architecture}
In DDPG, each robot requires two sets of neural networks: i) actor network and ii) critic network.
The actor network consisted of 3 fully connected layers with 512, 512 parameters followed by a rectified linear activation function (ReLU) nonlinearities between the input and output layers. Outputs were connected to sigmoid and tanh activation functions to limit the range to ($0 \leq v \leq 1$ and $-1 \leq \omega \leq 1$). The range of $v$ and $w$ is further processed to limit the velocity applying the motion constraints of the robot as described with action space.
The critic network, consisted of input and output layers, which were observations and state-action values, and 2 fully connected layers with the same number of parameters as the actor network. Unlike the actor network, after the first layer, actions were concatenated and fed into the second layer. 

\begin{table}[t]
\centering
\caption{Parameter Values for DRL training}
\begin{tabular}{lr}
\toprule
\multicolumn{1}{c}{Parameters $\qquad \qquad \qquad \qquad \qquad \qquad \qquad$} &  \multicolumn{1}{c}{Values}  
\\ \toprule
Total number of episodes, $M$ & 120 \\
Training steps per episode, $T$ & 1024\\
Total number of robots, $N$ & 4 \\
Discount factor, $\gamma$ & 0.99 \\
Soft weight update factor, $\tau$ & 0.5 \\
Goal reward, $R_g$ & 100.0 \\
Collision penalty, $R_c$ & -100.0 \\
Progress reward factor, $a$ & 4.0 \\
Approaching penalty parameter, $\lambda$ & $\log{2}$ \\
\toprule
\end{tabular}
\label{table:parameters}
\end{table}

\subsection{Metrics}

\subsubsection{Training Performance Metrics}
To evaluate the training
performance of each training strategy, we measured three
training performance metrics are below.

 \begin{itemize}
    \item Average reward, $r_{avg}$,  is the averaged value of the re-
ward obtained for one episode over four agents in the
experiment.
    \item Catastrophic interference, $N_{ci}$, is the number of the events
when the average reward changes by more than 50\% of the
the range between the maximum and minimum values of
average reward over the training. 
    \item Failed agent, $N_{fa}$, represents the average number of
failed agents during one training instance. The training
agents are regarded as failed agents when the difference
between the average reward at the beginning and the end
during training is within 1, and the difference between
the maximum and minimum value is under 1.5.
\end{itemize}

\subsubsection{Evaluation Performance Metrics}
 To assess the performance of trained models with IDDPG, SNDDPG, SEDDPG and FLDDPG in the evaluation stage, we defined two performance metrics: i) mission success rate and ii) mission completion time
 
 \begin{itemize}
    \item Success rate, $\rho_s$, is the rate of the successful episode from all the episodes without collision within one episode.
    \item Completion time, $t_{comp}$, represents the average time taken for all the robots to reach the targets without collision and within the time limit.
\end{itemize}

Success rate shows the robustness of the controller and completion time represents the optimisation performance of each training strategy. In the simulated evaluation experiment, the success rate and completion time were calculated by averaging performance of the four agents in the environment over 20 runs with four trained models for each agent during the training stage.

\subsection{Real-Robot Experiment}
\begin{figure}[t]
    \centering
    \includegraphics[width=0.48\textwidth]{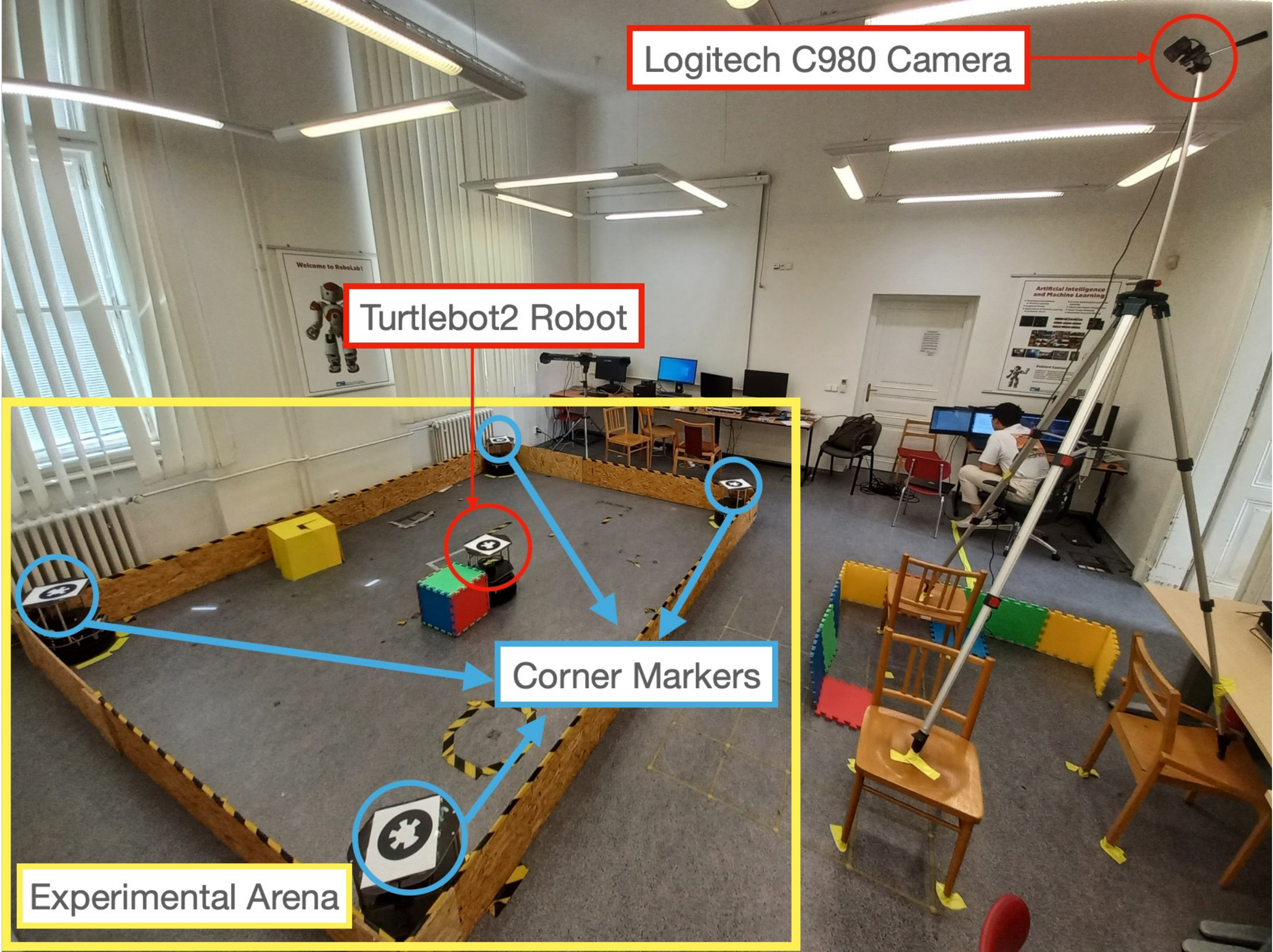}
    \caption{ A snapshot of the experimental arena.
    The turtlebot2 mobile robot performs navigation and collision avoidance in the rectangular arena. The overhead camera is used for real-time localisation of the robot in the two-dimensional arena coordinate frame, which is determined by the four corner markers.}
    \label{fig:experimental_arena}
\end{figure}

Experiments including real-robots were performed in ${4\times3m^2}$ arena with one robot. The arena had several obstacles
similar to the ones in the simulation, such as boxes. The
camera is mounted outside the arena, and the main PC is connected
to the camera for real-time localisation. The snapshot of
the experimental arena is shown in Fig.~\ref{fig:experimental_arena}.
 
\begin{figure}[t]
    \centering
    \includegraphics[width=0.24\textwidth]{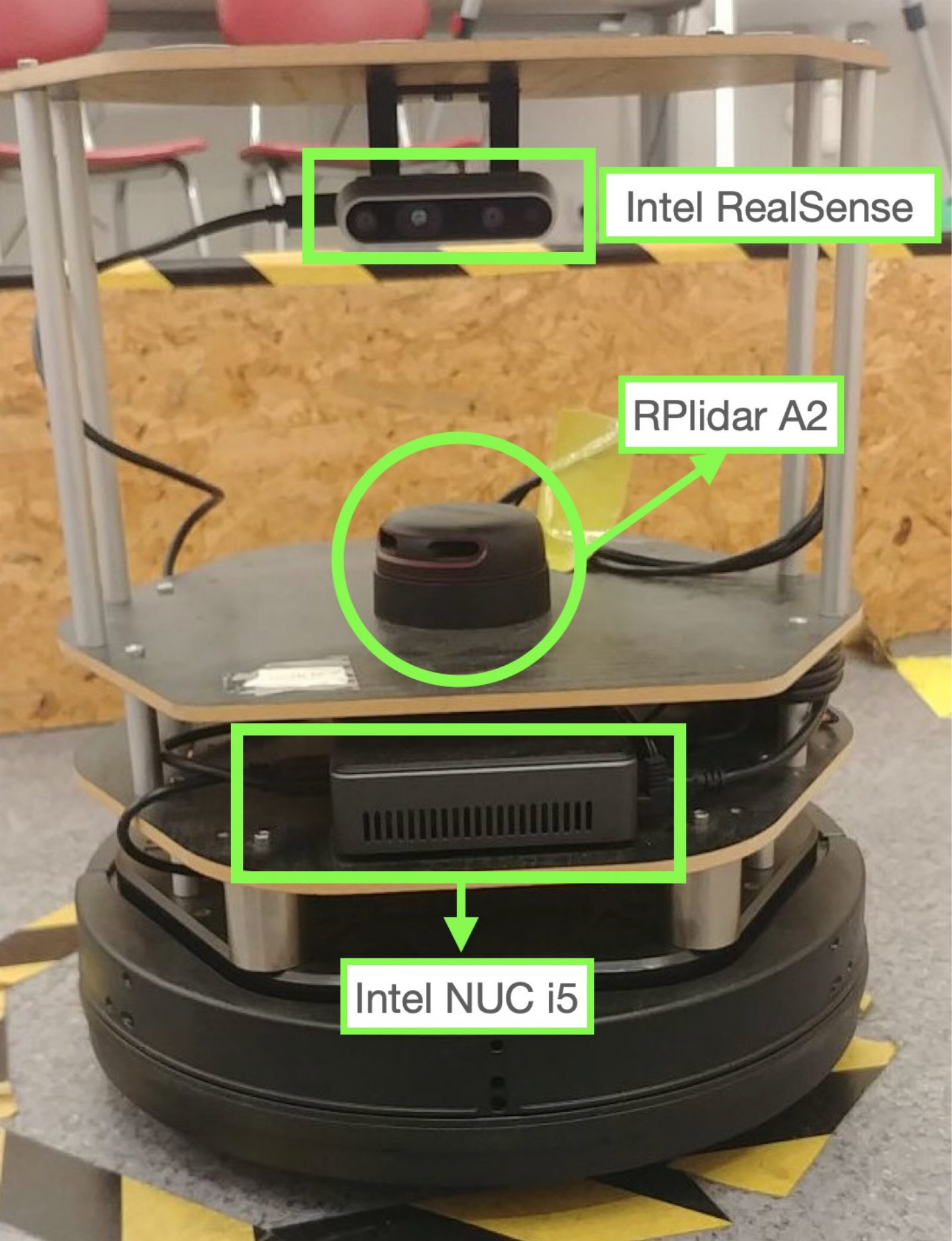}
    \caption{Single Turtlebot2 from top to bottom with a camera, RPlidar, Intel NUC and base platform.}
    \label{fig:turtle}
\end{figure}
\subsubsection{Hardware Specification}
 Turtlebot2 mobile robot was
used as a robotic platform, depicted
in Fig.~\ref{fig:turtle}. The Turtlebot2 is a tower-shape differential drive mobile robot with a diameter
of~38~cm and height of 60~cm. The robots use a differential
drive with a coaster wheel and can turn on the spot
with a maximum rotational speed of ~3 rad/s and translational speed of ~0.65m/s. 
Robots were equipped with an RPlidar A2
which is planar LiDAR mounted parallel to the ground with
a 10~m maximum range (covering the entire arena) with 720 beams per 360\textdegree @ 10~Hz. The minimum range is 0.2~m, within the robot's footprint. Robots are also equipped with Intel
RealSense camera. However, the camera is not used in the
experiments. For computation, the robot is mounted with Intel
NUC i5 of 8th generation equipped with Ubuntu 20.04 and
ROS Noetic.

\subsubsection{System Specification}
Master PC with AMD Threadripper 3960X, Nvidia RTX 3090 and 64GB RAM was used as a centralised hub to which all the robots are connected via Wi-Fi. 
The master PC was used for inter-robot communication and training neural network models. 
Additionally, this PC provides a position for all the robots running relevant ROS nodes for the tracking system.

\subsubsection{Tracking System}
To provide each robot with its position and orientation in the arena, the Logitech C980 camera was used with 1080p~@30~Hz resolution to detect and localise fiducial markers placed on top of the robot.
The WhyCode fiducial markers~\cite{ulrich2022towards, lightbody2017efficient} were used for the external localisation of the robots.
The WhyCode is a low-cost vision-based localisation system capable of real-time pose estimation of extensive number of black-and-white circular fiducial markers.
It is capable of unique identification and 6 degree-of-freedom (DOF) pose estimation with high precision using only an off-the-shelf web camera.
All this was computed and then transferred to each robot from the master PC via Wi-Fi. The 6~DOF estimated information was used as state information for the trained neural network to infer desirable actions for the task.

%% file: src/results.tex
\section{RESULTS \& DISCUSSION}

\begin{figure}[t]
    \centering
    \includegraphics[width=0.40\textwidth]{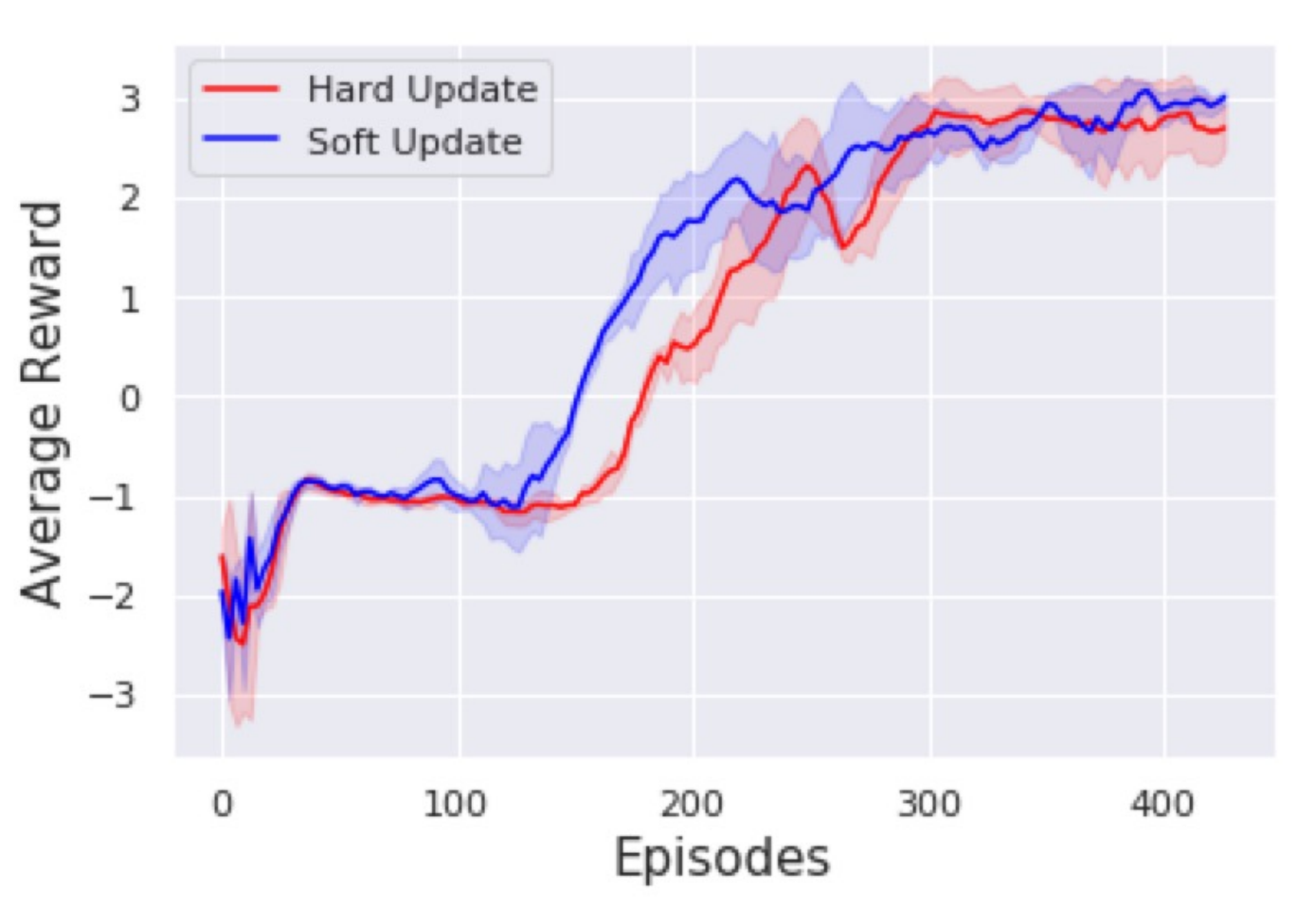}
    
    \caption{Average reward per episode of two different weight averaging methods: (red line) hard update and (blue line) soft update.}
    \label{fig:weight_averaging}
\end{figure}

\subsection{Results of FLDDPG Design Experiments}
In this section, the results from experiments to evaluate the two weight averaging methods: i) hard update and ii) soft update were reported and discussed. This experiment uses the same experimental scenario of collision avoidance and navigation. However, it is conducted with the smaller number of time steps per episode and different reward design than the main experiment in a fast manner to evaluate the performance of hard update and soft update only. Fig.~\ref{fig:weight_averaging} shows the results from the experiments testing training performance of FLDDPG with hard update and soft update. 

In Fig.~\ref{fig:weight_averaging}, it is illustrated that the average reward started to increase earlier and converged faster with the soft update method than the hard update method. 
We found that the performance of each robot temporarily decreases after weight update with the averaged model using the hard update method. Since the averaged model generalises individual local models, its performance in each local environment varies. The result shows that the soft update method prevented the adverse conversion from the local model to the averaged model, resulting in faster training by 18\% reducing training time compared to the hard weight update.

\begin{figure}
    \centering
    \includegraphics[width=0.48\textwidth]{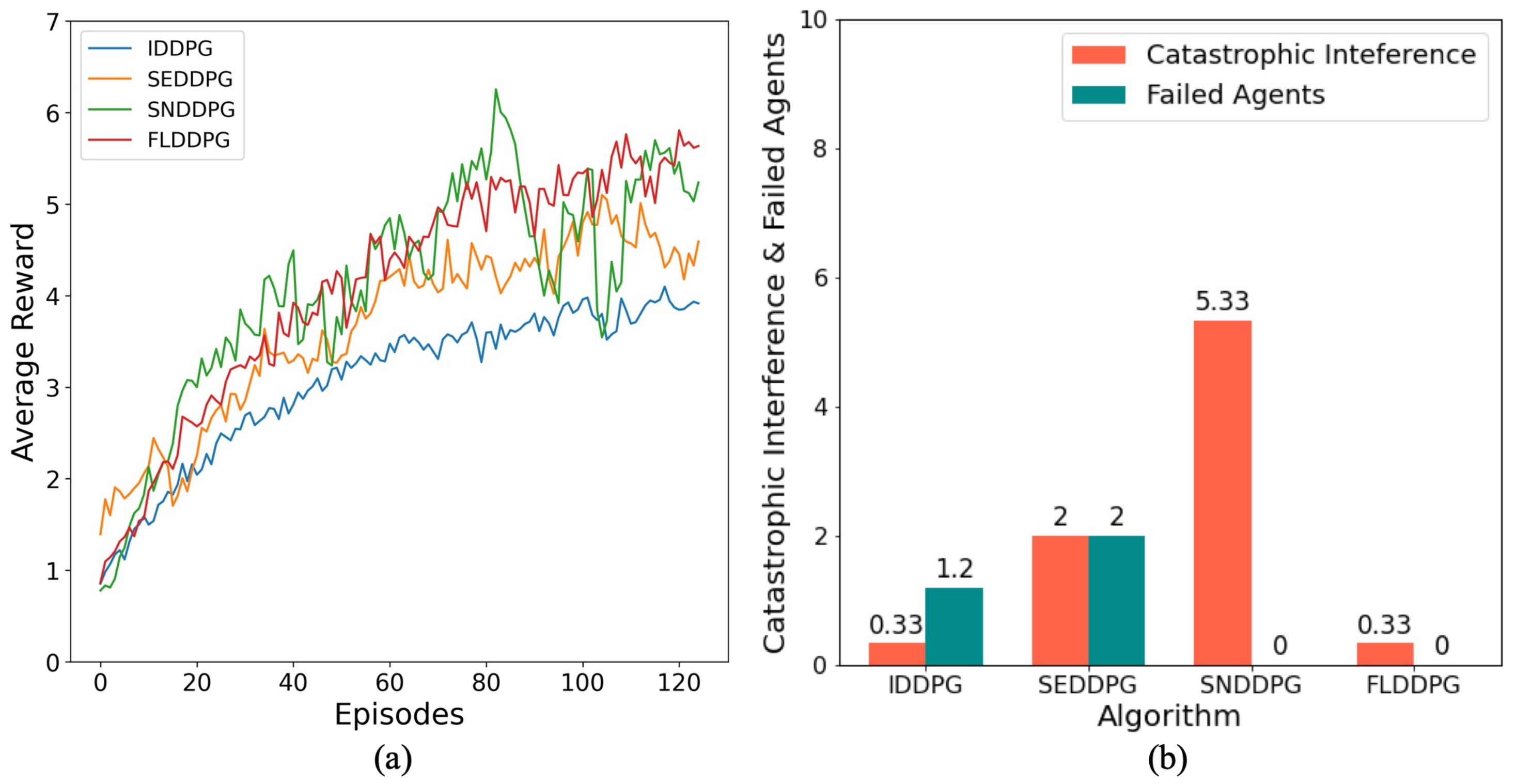}
    \caption{The training results of four different algorithms: IDDPG, SNDDPG, SEDDPG and FLDDPG. (a) Average reward per episode over training. (b) Average number of catastrophic interference and failed agents per training.}
    \label{fig:training_result_ci_fa}
\end{figure}




\subsection{Training Performance of the Four Strategies}

During the training, the robots learned navigation and collision avoidance in the environment, illustrated in Figure~\ref{fig:experiment_environment} with four training strategies. As described in the last section, each strategy applies the limitation of the total transferred data volume. The result from the training is illustrated in Fig.~\ref{fig:training_result_ci_fa}. Fig.~\ref{fig:training_result_ci_fa}~(a) shows the average rewards of each training strategy across the four agents over training instances with three different random seeds, and Fig.~\ref{fig:training_result_ci_fa}~(b) show the average number of catastrophic interference and the number of failed agents over three independent runs.

In Fig.~\ref{fig:training_result_ci_fa}~(a), it is shown that the average reward of IDDPG is the lowest and followed by SEDDPG, SNDDPG and FLDDPG in ascending order at the end of training. This result means the overall training performance of the four agents is the lowest with IDDPG and the highest with FLDDPG. The performance of the trained models with different training strategies is more investigated with the evaluation experiments, described in Section IV-C and IV-D. One interesting observation in Fig.~\ref{fig:training_result_ci_fa}~(a) is that the average reward of SNDDPG fluctuates with greater amplitudes than other strategies. This observation about fluctuation is more quantitatively evaluated in Fig.~\ref{fig:training_result_ci_fa}~(b).

Fig.~\ref{fig:training_result_ci_fa}~(b) shows the number of catastrophic interference and failed agents per training instance. For the number of catastrophic interference, IDDPG and FLDDPG had the lowest value with 0.33 per training instance. SEDDPG resulted in a slightly higher value of 2. With SNDDPG, the number of catastrophic interference had the highest value with 5.33, which explains the dramatic fluctuation observed in Fig.~\ref{fig:training_result_ci_fa}~(a). For IDDPG, as the agents are trained independently, it is not affected by updated shared experience replay buffer or shared network update. Therefore, it resulted in a low number of catastrophic interference. For SEDDPG, as the experience replay buffer is updated with large intervals, and the data samples for each agent can drastically change. Also, during the intervals, individual networks are trained in a personalised way with the individual data, and the update of the experience replay buffer with the shared aggregate data can deteriorate the training of the models. This drastic change resulted in a higher number of catastrophic interference compared to IDDPG. For SNDDPG, during the intervals between updates, the agents collect samples with the fixed models. The fixed models can cause collecting only trivial data, which does not improve the model's improvement when updated. Subsequently, the trained model with trivial data can cause the collection of other trivial data, forming a vicious cycle. Moreover, when the shared network is updated using the collected samples at large intervals, there could be a drastic change in the shared network. The collected samples with the non-improving model can store non-beneficial samples for network improvement.


For the number of failed agents, both SNDDPG and FLDDPG had no failed agents. However, IDDPG and SEDDPG resulted in 1.2 and 2 average number of failed agents per training instance,  respectively. For IDDPG, there is no data exchange or model exchange mechanism so that one agent cannot benefit from other agents. Their failure happens as they do not have a sample or a neural network model generating samples that are beneficial to update a neural network model in an improving way. For SEDDPG, as their data transfer is limited due to the limitation on the total transferred data volume, the experience replay buffers are updated every 5 episodes. It can be presumed that this infrequent update of the experience replay buffer caused noise in the samples, thereby resulting in a higher number of failed agents. On the other hand, SNDDPG and FLDDPG contribute to individual agents getting benefits from the other agents by sharing samples to update the shared network and neural network parameters.

\subsection{Simulation Evaluation Results}
The trained models for each agent using four different strategies were evaluated by running the task in the same simulated environment where the training is done.  For each
model, the task was performed 20 times, and the results 
averaged over 20 runs and four agents.

\begin{figure}[t]
    \centering
    \includegraphics[width=0.46\textwidth]{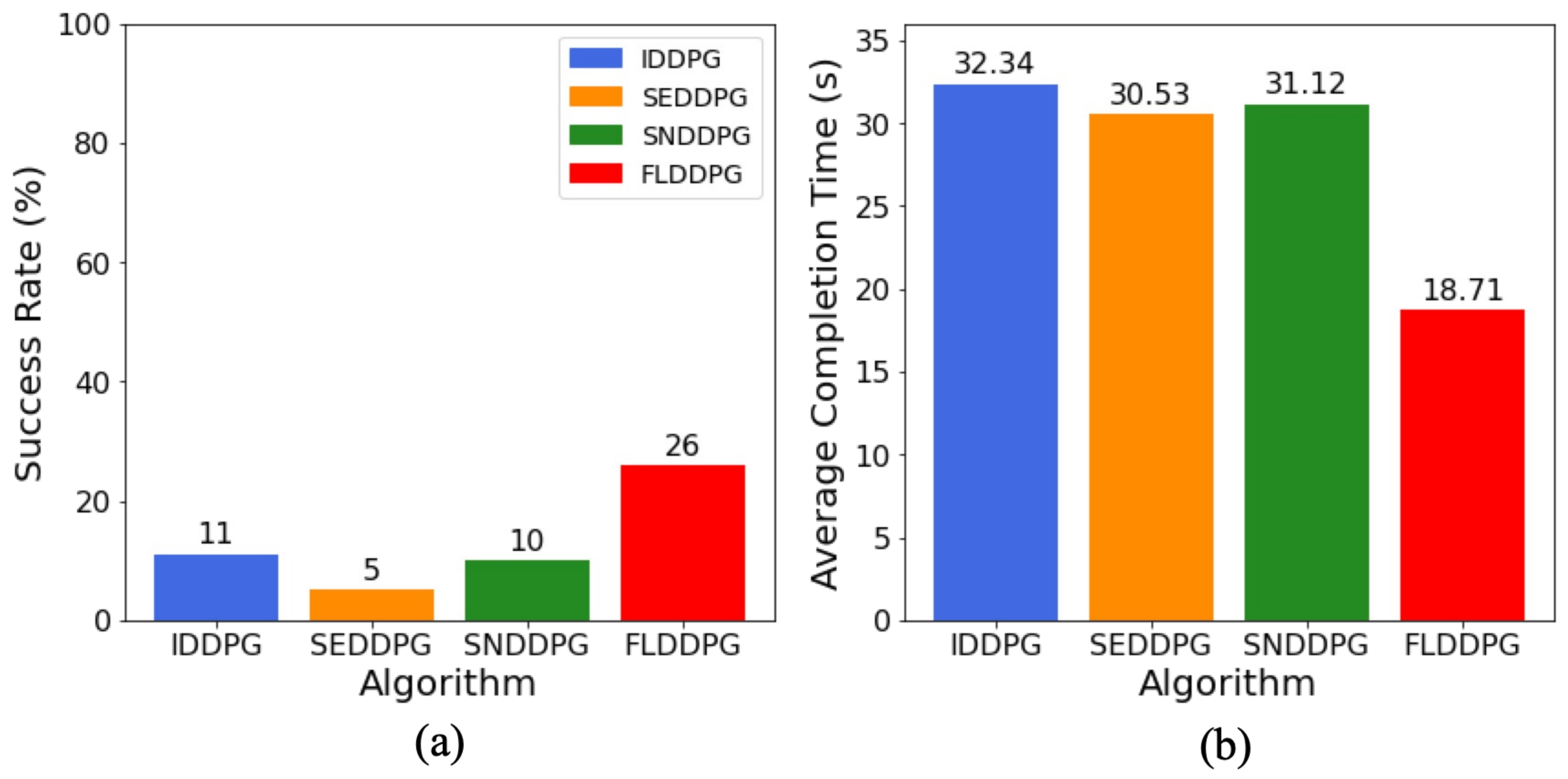}
    \caption{The evaluation experiment results with the four different DRL training strategies. (a) and (b) show success rate and average completion time of 4 agents over 20 runs of the four algorithms respectively.}
    \label{fig:evaluation_results}
\end{figure}

Fig.~\ref{fig:evaluation_results} illustrates the evaluation results in the simulated environment. Fig.~\ref{fig:evaluation_results}~(a) shows the success rates and (b) shows the average completion time. In Fig.~\ref{fig:evaluation_results}~(a), it is shown that IDDPG, SEDDPG, SNDDPG and FLDDPG scored 11\%, 5\%, 10\% and 26\% success rates, respectively. By comparing the scores, it is shown that the average performance of models trained with SEDDPG is the lowest. This can be explained by the number of catastrophic interference and failed agents observed in the training performance (shown in Fig.~\ref{fig:training_result_ci_fa}). Since the trained models from the failed agents never succeeded in the task, it dramatically influenced the low success rate of SEDDPG. Furthermore, the high number of catastrophic interference also negatively influenced the performance of the trained models, thereby causing a low success rate. 

The effects of the number of catastrophic interference and failed agents are further shown with the success rates of SNDDPG and IDDPG, respectively. SNDDPG resulted in the greatest number of catastrophic interference while having no failed agents and IDDPG resulted in the comparably high number of failed agents while having the lowest number of catastrophic interference. Both SNDDPG and IDDPG scored 10\% and 11\%, which are considerably lower than the success rate of FLDDPG (26\%).

Compared to the three baselines, FLDDPG scored a higher success rate of 26\%, which is approximately 2.36 times improvement than the second best method (IDDPG, 11\%). This reason for this improvement can be supported by the significantly low number of catastrophic interference (0.33) and the absence of the failed agent. Furthermore, FLDDPG achieved a considerably short average completion time than other baselines, which is only approximately 60\% of the second short average completion time (SEDDPG, 30.53~s). Both high success rate and short average completion time manifest the robustness and generalisation of FLDDPG.

\begin{figure}[t]
    \centering
    \includegraphics[width=0.49\textwidth]{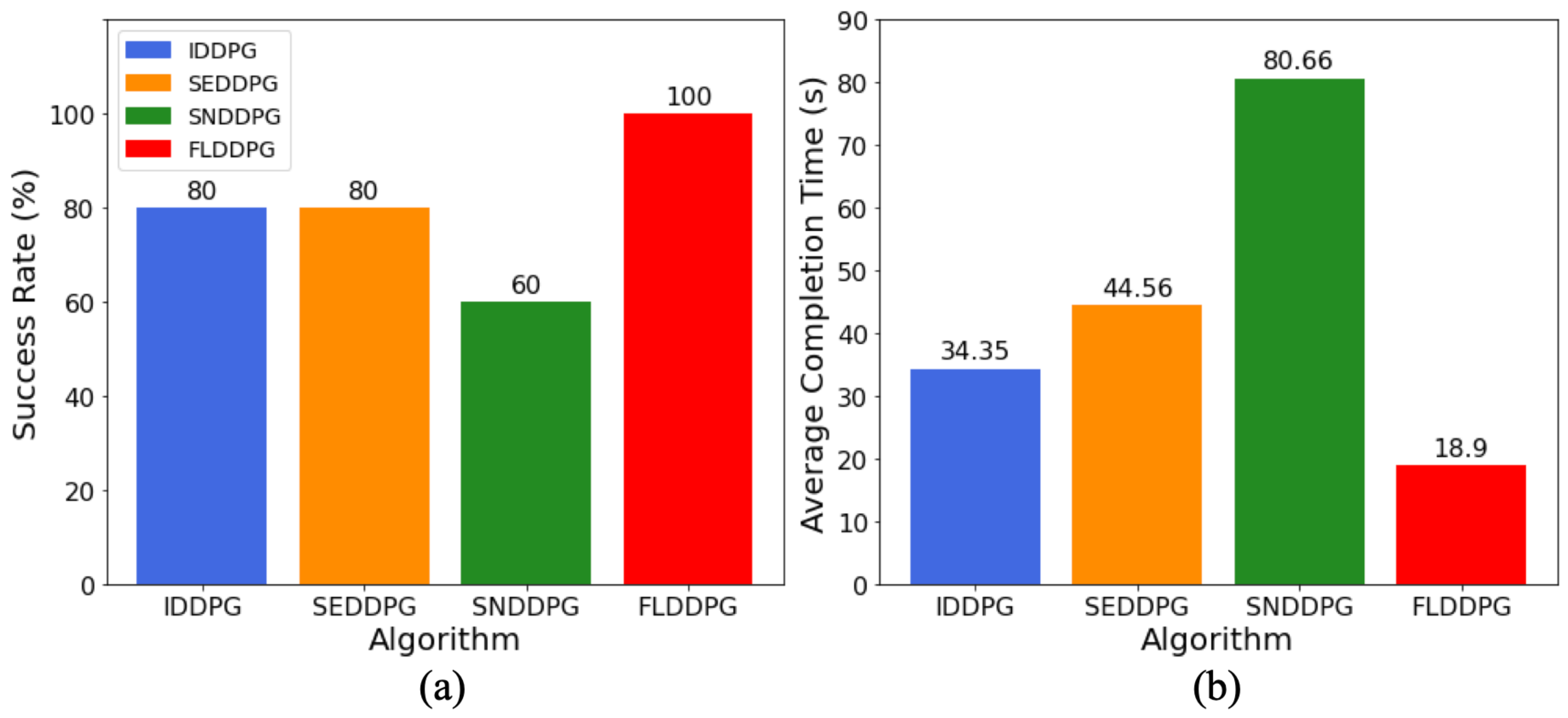}
    \caption{The real robot experiment results with the four different DRL training strategies. (a) and (b) show success rate and average completion time over 5 runs of the four algorithms respectively using the most successful model.}
    \label{fig:real_robot_results}
\end{figure}

\subsection{Real-Robot Experiment Results}

For the real-robot experiments, the most successful trained model for each strategy under an identical communication period constraints are chosen and used for the real-robot experiments. Fig~\ref{fig:real_robot_results}~(a) and (b) illustrate the success rate and average completion time of four training strategies in the real-robot experiment. 

The result shows that the FLDDPG achieved the highest success rate and lowest average completion time. This supports the finding from the evaluation experiments in simulation that FLDDPG showed robustness and generalisation ability when it is transferred into the real-robot platform.

\subsection{Discussion}
From the three sets of results, i) training performance, ii) evaluation in simulation and iii) evaluation with a real robot, it is found that the proposed FLDDPG strategy prevented failure of training of individuals in a collective learning scenario by aggregating the individual neural networks periodically compared to the independent learning scenario (IDDPG). Moreover, under the limitation of the communication bandwidth, FLDDPG showed greater robustness and generalisation ability by evaluation in the simulation and the real robot setting. This finding from the experimental results about FLDDPG suggests that FLDDPG can benefit the autonomous swarm robotic systems operating in uncertain environments where online learning is required under the limitation of communication bandwidth. 

The examples of such environments are underground mines and tunnels, where swarm robotic systems are required to perform inspection or search-and-rescue missions. 
The teams participating in the Defense Advanced Research Projects Agency (DARPA) subterranean challenge actually reported that one of the most difficult challenges was the limited communication bandwidth \cite{Roucek2020DARPA}, preventing the robots from efficiently exchange sensorimotor data.
Thus, we believe that the use of FLDDPG in similar scenarios would improve the robustness of the deployed multi- and swarm-robotic systems.


%% file: src/conclusion.tex
\section{CONCLUSION}
In this paper, we proposed a federated learning-based deep reinforcement learning training strategy for swarm robotic systems, FLDDPG.
FLDDPG reduces their reliance on high-fidelity communications and protects confidentiality of the locally collected data.
Through the experiments in a collective learning scenario for navigation and collision avoidance under the limitation of communication bandwidth, FLDDPG showed higher robustness and better generalisation compared to the three baselines, IDDPG, SEDDPG and SNDDPG.
This finding is supported by both simulation and real-robot experiments.
The result suggests that FLDDPG can benefit autonomous swarm robotic systems operating in environments with the limited communication such as underground or underwater.
In the future, the federated learning based DRL training strategy with heterogeneous swarm robotic systems in the real, changing environments will be developed and investigated.
